\documentclass[letterpaper, 10 pt, conference]{ieeeconf}  % Comment this line out if you need a4paper

\IEEEoverridecommandlockouts                              % This command is only needed if 
\usepackage{graphics} % for pdf, bitmapped graphics files
\usepackage{epsfig} % for postscript graphics files
\usepackage{mathptmx} % assumes new font selection scheme installed
\usepackage{times} % assumes new font selection scheme installed
\usepackage{amsmath} % assumes amsmath package installed
\usepackage{amssymb}  % assumes amsmath package installed
\usepackage{cite}
\usepackage{color}
\usepackage{multirow}
\usepackage[table,xcdraw]{xcolor}
\usepackage{hyperref}

\usepackage{xcolor}
\usepackage{array}
\usepackage[table]{xcolor}
\usepackage{comment}
\usepackage{float}
\usepackage{graphicx}
\usepackage{caption}
\usepackage[skip=0cm,list=true,labelfont=it]{subcaption}
\usepackage{xspace}
\usepackage{multirow}
\usepackage{vcell}
\newcommand\blfootnote[1]{%
  \begingroup
  \renewcommand\thefootnote{}\footnote{#1}%
  \addtocounter{footnote}{-1}%
  \endgroup
}
\begin{document}
\title{\LARGE \bf
ADOD: Adaptive Domain-Aware Object Detection with Residual Attention for Underwater Environments
}

\author{Lyes Saad Saoud$^{*}$,  Zhenwei Niu, Atif Sultan, Lakmal Seneviratne and Irfan Hussain$^{**}$% <-this % stops a space
\thanks{}% <-this % stops a space
\thanks{$^{1}$Mechanical Engineering Department, Khalifa University, Abu Dhabi, United Arab Emirates, P O Box 127788, Abu Dhabi, UAE}%
\thanks{$^{2}$Khalifa University Center for Autonomous and Robotic Systems, Khalifa University, Abu Dhabi, United Arab Emirates, P O Box 127788, Abu Dhabi, UAE.$^{**}$ Email: lyes.saoud@ku.ac.ae}%
\thanks{}% <-this % stops a space
\thanks{$^{**}$ Corresponding Author, Email: irfan.hussain@ku.ac.ae}%
% \thanks{$^{**}$ Email: irfan.hussain@ku.ac.ae}%
}

\maketitle
\thispagestyle{empty}
\pagestyle{empty}

%%%%%%%%%%%%%%%%%%%%%%%%%%%%%%%%%%%%%%%%%%%%%%%%%%%%%%%%%%%%%%%%%%%%%%%%%%%%%%%%

\begin{abstract}
This research presents ADOD, a novel approach
to address domain generalization in underwater object detection. Our method enhances the model’s ability to generalize
across diverse and unseen domains, ensuring robustness in
various underwater environments. The first key contribution
is Residual Attention YOLOv3, a novel variant of the YOLOv3
framework empowered by residual attention modules. These
modules enable the model to focus on informative features
while suppressing background noise, leading to improved detection accuracy and adaptability to different domains. The
second contribution is the attention-based domain classification
module, vital during training. This module helps the model
identify domain-specific information, facilitating the learning of
domain-invariant features. Consequently, ADOD can generalize
effectively to underwater environments with distinct visual
characteristics. Extensive experiments on diverse underwater
datasets demonstrate ADOD’s superior performance compared
to state-of-the-art domain generalization methods, particularly in challenging scenarios. The proposed model achieves
exceptional detection performance in both seen and unseen
domains, showcasing its effectiveness in handling domain shifts
in underwater object detection tasks. ADOD represents a
significant advancement in adaptive object detection, providing
a promising solution for real-world applications in underwater
environments. With the prevalence of domain shifts in such
settings, the model’s strong generalization ability becomes a
valuable asset for practical underwater surveillance and marine
research endeavors.ADOD’s code is available on GitHub at
https://github.com/LyesSaadSaoud/ADOD

% Robotic vision in underwater environments presents unique challenges that demand innovative solutions to enhance performance and reliability. In this paper, we introduce UW-Detect, an advanced approach for underwater object detection enriched with Multi-Scale Attention (MSA). UW-Detect is designed to address domain generalization challenges, ensuring robustness in diverse underwater environments.
% UW-Detect combines two significant contributions. First, it employs Residual Attention YOLOv3, a novel variant of the YOLOv3 framework empowered by residual attention modules. This enhances detection accuracy and adaptability by focusing on informative features while suppressing background noise. Second, the incorporation of Multi-Scale Attention (MSA) further refines feature representation by emphasizing informative regions and reducing noise.
% Extensive experiments on diverse underwater datasets demonstrate UW-Detect's superior performance compared to state-of-the-art domain generalization methods, especially in challenging scenarios. The model excels in both seen and unseen domains, showcasing its effectiveness in handling domain shifts in underwater object detection tasks.
% UW-Detect represents a significant leap in adaptive object detection, offering a promising solution for real-world applications in underwater robotics and marine research. Its robustness and generalization capabilities make it a valuable asset for practical underwater surveillance and exploration. UW-Detect's code is available on GitHub at https://github.com/LyesSaadSaoud/UW-Detect

\blfootnote{Paper presented at the 21st International Conference on Advanced Robotics (ICAR 2023)}
\end{abstract}

\begin{keywords}
Object Detection, Domain Generalization, Adaptive Learning, Residual Attention, Attention-Based Domain Classification, Underwater Scenes.
\end{keywords}
\begin{figure}
    \includegraphics[width=\columnwidth]{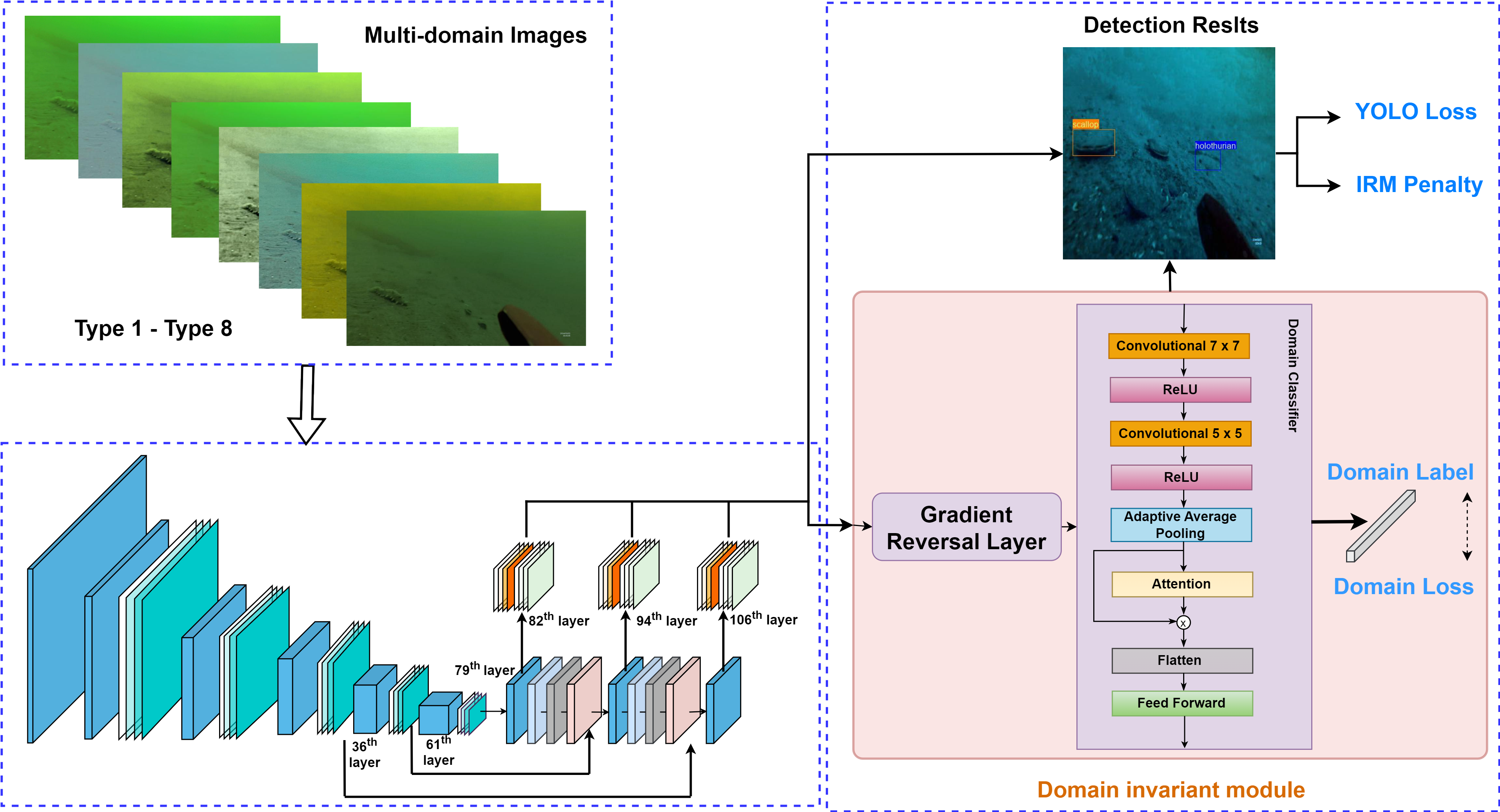}
    \caption{The ADOD model architecture for underwater object detection, an enhanced version of YOLOv3, incorporates residual layers and channel attention mechanisms to achieve accurate and efficient detection. The model utilizes a fully convolutional network with a Darknet-53 backbone and employs three feature maps of varying sizes for object detection. Residual layers are strategically placed after specific feature map layers (e.g., $82^{nd}$, $94^{th}$, and $106^{th}$) to address gradient challenges during training, while multi-scale attention modules enhance feature representation by focusing on informative regions and discarding noise.}
    \label{fig:Image1}
\end{figure}
%%%%%%%%%%%%%%%%%%%%%%%%%%%%%%%%%%%%%%%%%%%%%%%%%%%%%%%%%%%%%%%%%%%%%%%%%%%%%%%%
\section{Introduction}

Underwater object detection is of utmost importance for ensuring safe maritime navigation and preserving the environment. It plays a vital role in identifying submerged hazards like rocks, wrecks, and coral reefs, assisting vessels in navigating treacherous waters and avoiding collisions. Additionally, underwater object detection is crucial for locating and maintaining underwater infrastructure, including pipelines, cables, and offshore platforms, contributing to damage prevention and effective infrastructure management. Moreover, it facilitates the identification and removal of underwater debris and pollutants, safeguarding marine life.

Despite its significance, underwater object detection faces challenges due to inherent variations in object appearance. Factors such as viewpoint, background, lighting conditions, and image quality introduce complexity and diversity in visually representing objects.  Light interference underwater and the effects of color casts that the main challenges when dealing with images underwater\cite{FU2023243}.  In addition to these factors low contrast and blur in underwater images make it difficult for the image processing algorithm to reproduce the results similar to trained data-set.\cite{XU2023204}

Consequently, Domain shift occurs when the training data's distribution doesn't match the testing data, leading to a drop in model performance in real-world scenarios \cite{recht2019imagenet-1, hendrycks2019benchmarking-2}. The challenges in underwater object detection aren't just about image quality. Creating a dataset for this task is tough. It involves expensive and demanding processes like collecting images with divers or underwater robots, which require substantial preparation and funding to build a limited dataset. These images also need expert annotations, which are costly and time-consuming. This whole process limits the data available for training underwater object detection algorithms \cite{liu2020towards-8} and makes this approach less practical due to a scarcity of labeled datasets.

Dealing with domain shift by diversifying and increasing training data is hindered by the labor-intensive image annotation process. Domain adaptation techniques adjust model parameters using new data from the target domain, but getting enough suitable target data remains a challenge \cite{chen2018domain-3, xu2020exploring-4, hsu2020progressive-5}.

Recent progress in domain generalization offers an alternative to address domain shift. Unlike domain adaptation, it aims to create a set of model parameters that work well in various new domains without needing data from those domains or extra adjustments \cite{li2017deeper-6}. This flexibility is valuable as it allows models to perform effectively in diverse and unexpected conditions, even with limited training data. Various techniques enhance domain generalization. For example, Ji-Gen uses a jigsaw puzzle approach\cite{carlucci2019domain-7}, while another method aligns domain distributions using autoencoders and Maximum Mean Discrepancy \cite{li2018domain-10}. Episodic training \cite{li2019episodic-11} toughens models by exposing them to domain shifts, and data augmentation techniques, like CrossGrad \cite{shankar2018generalizing-12}, enhance model generalization through adversarial modifications.

In underwater settings, domain generalization techniques like DG-YOLO \cite{liu2020towards-8} have proven valuable. They create detectors that work well in different underwater scenarios by combining domain adversarial training and invariant risk minimization. This is important for applications like computer vision in autonomous driving and surveillance systems. Nevertheless, domain generalization in object detection, especially in challenging environments like underwater scenes, remains tough due to differences in lighting, visibility, and object appearances across domains.

To address these challenges, we propose ADOD, a novel Adaptive Domain-Aware Object Detection framework (Fig. \ref{fig:Image1}). At the core of ADOD is the Residual Attention Block, which efficiently captures and emphasizes essential spatial and channel-wise features. This empowers the model to focus on critical object characteristics and enhance detection accuracy.

Our contributions can be summarized as follows:
\begin{itemize}

\item Residual Attention Block: ADOD introduces a novel Residual Attention Block, combining the advantages of residual connections and channel attention mechanisms. This unique fusion enhances the model's ability to capture intricate object details and emphasize relevant features, leading to improved detection accuracy.

\item Domain Generalization: ADOD addresses the domain generalization problem by learning domain-invariant representations through the Residual Attention Block. This feature allows the model to seamlessly adapt to various domains, making it highly effective in detecting objects across diverse underwater conditions.

\item Adaptive Domain-Aware Object Detection: By incorporating attention-based domain classification, ADOD becomes sensitive to domain-specific information during object detection. This adaptivity enhances the model's ability to distinguish objects from different domains, resulting in superior detection performance.

\end{itemize}

\section{Related Works}

Numerous techniques have been proposed to enhance underwater object detection by addressing domain shift. For instance, RoIMix applies Mixup on the RoI level to simulate occlusion conditions, while dilated convolutions have been explored to improve feature extraction in underwater environments \cite{lin2020roimix, fan2020dual, chen2020c}. Attention-based and feature-pyramid-based methods also contribute to feature extraction; however, they may not fully resolve domain shift issues \cite{liang2022excavating, zhao2021composited}.

DG-YOLO \cite{liu2020towards} is an innovative method for addressing domain shifts in underwater object detection. It effectively handles limited data and domain diversity by using Water Quality Transfer as an augmentation technique. This approach combines domain adversarial training and invariant risk minimization to achieve domain-invariant detection performance.

%DG-YOLO emerged as one of the pioneering approaches to tackle underwater domain shift by introducing underwater object detection for domain generalization \cite{liu2020towards}. Leveraging a small dataset with limited domains, DG-YOLO employed Water Quality Transfer as an augmentation technique to expand dataset size and domain diversity, achieving domain-invariant detection performance through domain adversarial training and invariant risk minimization.

Domain Adaptation and Generalization represent crucial tasks in achieving strong performance in a target domain using source domain data. In domain adaptation, related source and target domains share the same label space, and techniques like DANN align features through adversarial training \cite{ganin2015unsupervised}. Other approaches generate target domain data from source domain data using VAE and GAN approaches \cite{xu2019adversarial, gong2019dlow}, with extensions to the detection task \cite{chen2018domain}.

On the other hand, domain generalization presents unique challenges as it involves training on multiple source domains and evaluating on an unseen but related domain. Various strategies have been proposed, including feature alignment between source domains, data augmentation-based methods for domain diversity, self-supervised training, aggregation-based methods, and meta-learning paradigms \cite{motiian2017unified, shankar2018generalizing, dou2019domain, d2018domain, balaji2018metareg, li2019episodic}. Despite significant progress in domain generalization, its application to underwater object detection remains relatively unexplored \cite{huang2019faster, lin2020roimix, liu2020towards}.Experiments with domain generalization  for underwater object detection show promising results on databases such VLCS, PACS, and S-UODAC2020\cite{CHEN202320}. 
%Domain generalization techniques applied on application like surgical tools detection had profound implications resulting in an method simplifies the training process as compared to standard\cite{Reiter2023}. Applications in field of Biomedical engineering include domain generalization for gastrointestinal tract images\cite{fan2022invnorm}. 
In this study, we introduce ADOD, an innovative Adaptive Domain-Aware Object Detection framework. ADOD harnesses the power of Residual Attention and attention-based domain classification to overcome domain generalization challenges. At the core of ADOD lies the novel Residual Attention Block, a component adept at capturing and emphasizing spatial and channel-wise features. This unique capability enables the model to focus on crucial object characteristics while suppressing irrelevant information, resulting in enhanced detection accuracy.

\begin{figure*}[ht]
\centering
\includegraphics[width=0.75\textwidth]{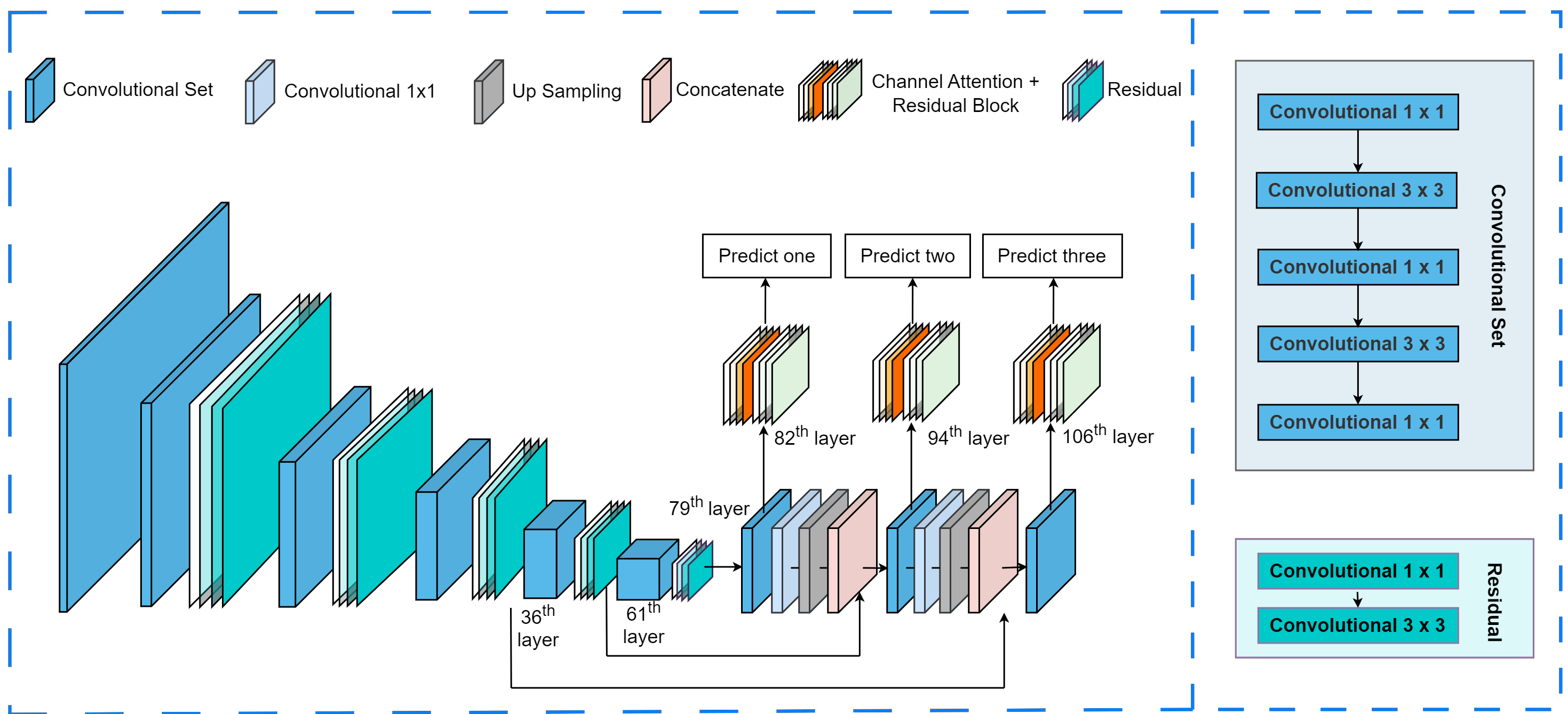}

\caption{An overview of the proposed ADOD model for underwater object detection. The enhanced YOLOv3-based architecture incorporates Residual Layers and Channel Attention modules, facilitating the accurate capture of intricate object details and emphasizing relevant features.}
\label{model_architecture}
\end{figure*}

\section{The proposed ADOD model}
YOLO (You Only Look Once) by Redmon et al. \cite{redmon2016you} is a groundbreaking deep learning model designed for real-time object detection. It integrates region proposal and classification stages into a unified architecture, prioritizing computational speed. 
%YOLOv2 and YOLOv3 further enhanced performance with anchor boxes and residual blocks. The YOLOv3 model's state-of-the-art accuracy and real-time capabilities make it widely adopted in computer vision applications.
The ADOD model represents a remarkable advancement in the field of underwater object detection, building upon the widely adopted YOLOv3 architecture \cite{redmon2018yolov3}. Tailored specifically for underwater scenarios, ADOD incorporates several pivotal modifications to elevate its performance, as illustrated in Figure \ref{model_architecture}.

At its core, ADOD is a fully convolutional network featuring a Darknet-53 backbone, distinguishing it from conventional models that typically include pooling layers. Instead, Darknet-53 employs four distinct types of residual units, each composed of a sequence of 1x1 and 3x3 convolutional layers. This innovative design enables the network to dynamically adjust the convolution kernel's stride during forward propagation, resulting in feature maps that are merely 1/25 of the original input size. Consequently, the tensor size is significantly reduced, making ADOD highly efficient.

For achieving precise object detection, ADOD leverages three feature maps of varying sizes. The Convolutional Set, deployed after the $79^{th}$ layer of the convolutional network, generates the detection result for the first scale (13x13). To capture more intricate details, the feature maps from the $79^{th}$ layer are upsampled and combined with the feature maps from the $61^{th}$ layer. Subsequently, they undergo further processing through convolutional layers, leading to the generation of the second scale feature map (26x26). Similarly, the third feature map is produced by upsampling and concatenating the output from the $36^{th}$ layer of the model.

To enhance the feature extraction process, two crucial components, Residual layers, and Channel Attention modules, are thoughtfully integrated into the ADOD model. These additions play a pivotal role in capturing intricate object details and adaptively emphasizing relevant features. As a result, ADOD demonstrates superior performance in underwater object detection, making it a highly effective solution for this challenging domain.

\subsubsection{Channel Attention}

In the ADOD model, the Channel Attention module plays a pivotal role in extracting features across different scales. By incorporating this module, ADOD significantly enhances its ability to capture finer details and emphasize relevant features, resulting in more accurate and robust object detection. The Channel Attention module is composed of several key components, including adaptive average pooling, two Conv2d 1x1 layers, one ReLU function, and one Sigmoid activation function, as depicted in Figure \ref{blocks_abc} (a). This combination of operations enables the module to intelligently focus on essential channel-wise information, effectively boosting the overall detection performance.

\subsubsection{Residual Layers}

\begin{figure*}[ht]
\centering
\includegraphics[width=0.75\textwidth]{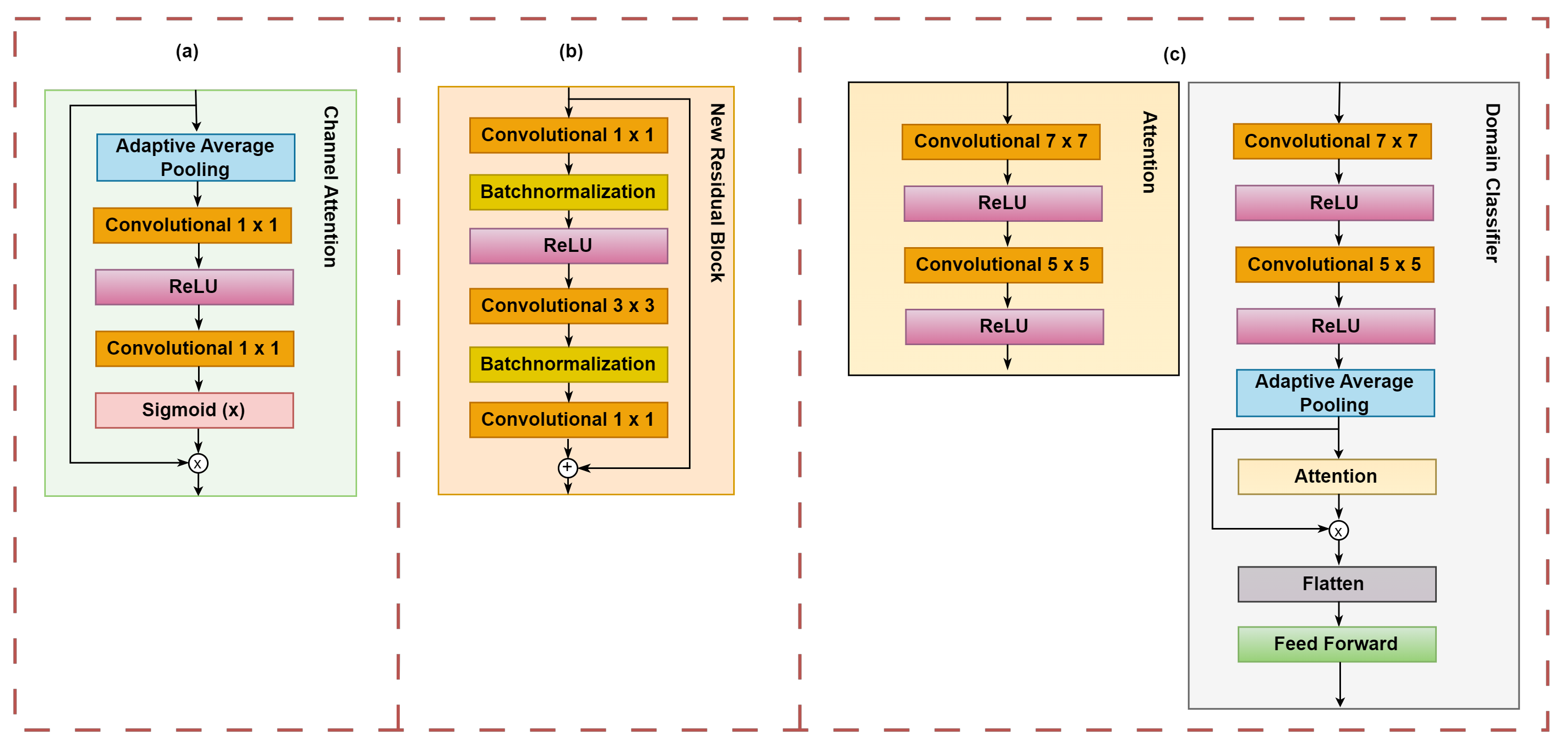}

\caption{New blocks introduced to YOLOv3. (a) Channel Attention block (b) Residual Block and (c) Domain Classifier}
\label{blocks_abc}
\end{figure*}

To tackle challenges related to gradient vanishing and explosion, which can hinder effective training in deep networks, ADOD introduces innovative Residual layers after each feature map layer (e.g., $82^{nd}$, $94^{th}$, and $106^{th}$ layers). As illustrated in Figure \ref{blocks_abc} (b), the Residual Block comprises two Conv2d 1x1 layers, one Conv2d 3x3 layer, two batch normalizations, and one ReLU activation function. These Residual layers act as shortcuts that allow the model to circumvent specific layers during forward propagation, facilitating smooth gradient flow and easing the optimization process. With the integration of Residual layers, ADOD can effectively train deeper networks, leading to superior model performance and more precise detection of underwater objects.

\subsubsection{Domain Classifier}

In the context of underwater feature detection and identification, the task becomes especially challenging due to the distortions introduced by the fluid medium, affecting both the objects and their surroundings. To enhance the algorithm's robustness in underwater conditions, the ADOD model incorporates a Domain Classifier module, aiming to improve the domain classification performance of YOLOv3 at each feature map layer.

The Domain Classifier, as depicted in Figure \ref{blocks_abc} (c), plays a crucial role in this modified version of YOLOv3. It consists of two conv2d layers (7x7 and 5x5 respectively), two ReLU functions, and an adaptive average pooling layer. The output of this process is then passed through an attention block, which consists of two conv2d layers (7x7 and 5x5 respectively) and two ReLU functions. The results obtained from the adaptive average pooling and attention block are multiplied and flattened before being fed into a feed forward block.

By introducing these new modules, including the Channel Attention, Residual layers, and Domain Classifier, to feature map layers, ADOD achieves superior object detection performance in harsh underwater conditions. As a result, it serves as a specialized tool specifically tailored for demanding underwater applications, effectively addressing the challenges posed by underwater environments.

\section{Results}
\begin{table*}[ht]
\caption{Detection results for different models using original and augmented datasets, with the values in
bold showing the best-obtained values.}
\label{tab:results}
\begin{tabular}{p{125pt}p{15pt}p{15pt}p{30pt}p{15pt}p{25pt}p{15pt}p{15pt}p{15pt}p{30pt}p{15pt}p{25pt}p{15pt}}
\hline
\multicolumn{1}{c}{}             & \multicolumn{6}{c}{Validation on original data}                                                                            & \multicolumn{6}{c}{Validation on augmented data (type8) }                                                           \\ \cline{2-13} 
\multicolumn{1}{c}{Method}       & echinus        & starfish       & holothurian    & scallop        & waterweed     & mAP           & echinus        & starfish       & holothurian    & scallop        & waterweed & mAP           \\ \hline
Baseline (YOLOv3)                & 83.67          & 71.87          & 51.32          & 64.54          & 0.00          & 54.28          & 69.08          & 20.18          & 31.20          & \textbf{44.58} & 0.00      & 33.01          \\
YOLO3+Residual                   & 80.28          & 64.24          & 47.31          & 50.78          & 0.00          & 49.12          & 63.95          & 32.50          & 32.52          & 40.10          & 0.00      & 33.81          \\
YOLO3+Channel attention          & 82.64          & 74.60          & 55.39          & 68.40          & 0.00          & 56.21          & 60.20          & 24.69          & 24.54          & 35.60          & 0.00      & 29.01          \\
YOLO3+Residual attention         & 82.96          & 74.51          & 54.10          & \textbf{69.12} & \textbf{4.76} & \textbf{57.09} & 66.31          & 27.39          & 32.46          & 34.13          & 0.00      & 32.06          \\
YOLO3+Domain                     & 82.04          & \textbf{77.66} & \textbf{59.86} & 62.10          & 0.00          & 56.33          & 58.97          & 31.61          & 28.30          & 37.24          & 0.00      & 31.22          \\
YOLO3+Domain +Residual           & 82.49          & 69.62          & 52.17          & 64.30          & 0.00          & 53.71          & \textbf{71.61} & 18.54          & 30.05          & 45.07          & 0.00      & 33.05          \\
YOLO3+Domain +Channel attention  & \textbf{84.88} & 74.95          & 52.95          & 67.61          & 0.68          & 56.21          & 65.21          & 28.35          & 30.12          & 41.79          & 0.00      & 33.09          \\
YOLO3+Domain +Residual attention & 82.72          & 69.96          & 52.15          & 65.61          & 0.00          & 54.09          & 67.97          & \textbf{38.91} & \textbf{34.28} & 38.33          & 0.00      & \textbf{35.90} \\ \hline
\end{tabular}

\end{table*}
\begin{figure*}
\begin{subfigure}{.12\textwidth}
   \includegraphics[width=\textwidth]{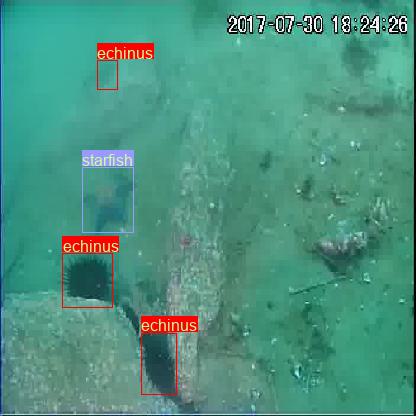}
   \caption{}
\end{subfigure}
% \hfill
\begin{subfigure}{.12\textwidth}
   \includegraphics[width=\textwidth]{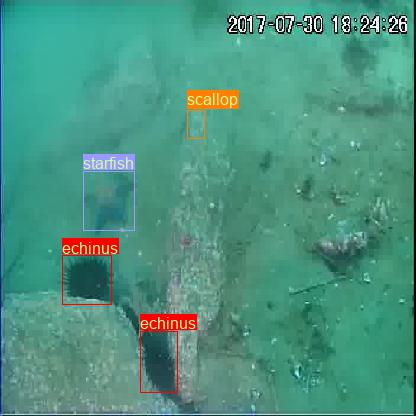}
   \caption{}
\end{subfigure}
% \hfill
\begin{subfigure}{.12\textwidth}
   \includegraphics[width=\textwidth]{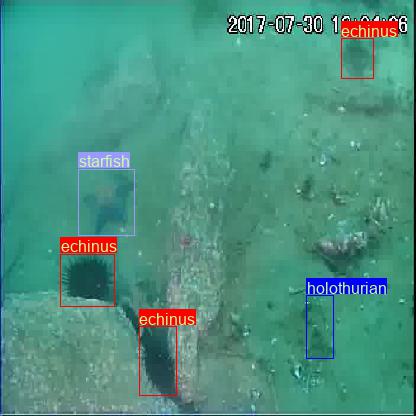}
   \caption{}
\end{subfigure}
% \hfill
\begin{subfigure}{.12\textwidth}
   \includegraphics[width=\textwidth]{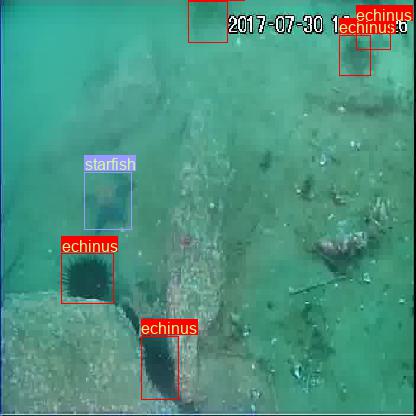}
   \caption{}
\end{subfigure}
% \hfill
\begin{subfigure}{.12\textwidth}
   \includegraphics[width=\textwidth]{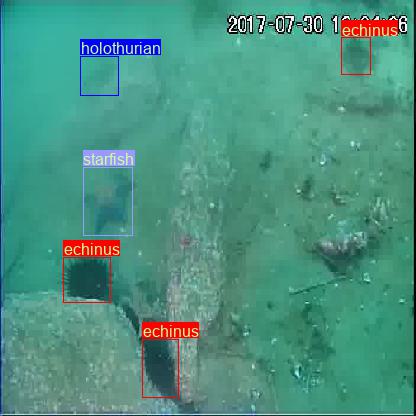}
   \caption{}
\end{subfigure}
% \hfill
\begin{subfigure}{.12\textwidth}
   \includegraphics[width=\textwidth]{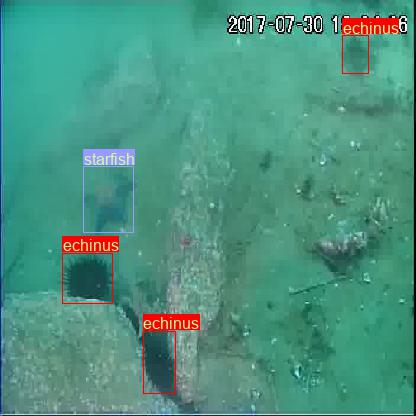}
   \caption{}
\end{subfigure}
% \hfill
\begin{subfigure}{.12\textwidth}
   \includegraphics[width=\textwidth]{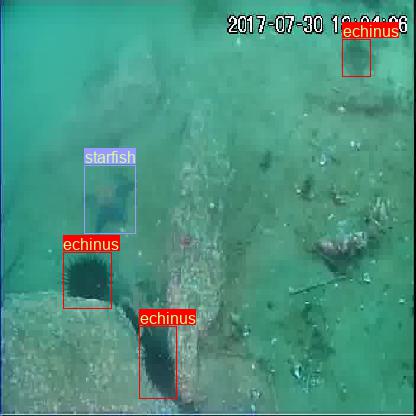}
   \caption{}
\end{subfigure}
% \hfill
\begin{subfigure}{.12\textwidth}
   \includegraphics[width=\textwidth]{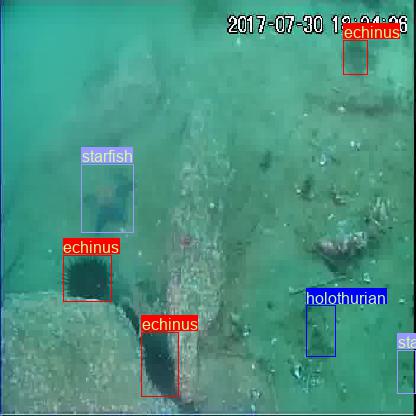}
   \caption{}
\end{subfigure}
\vfill
\begin{subfigure}{.12\textwidth}
   \includegraphics[width=\textwidth]{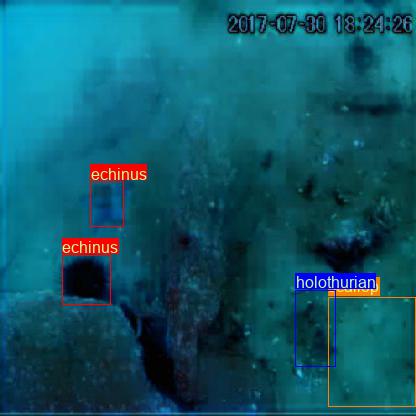}
   \caption{Model 1}
\end{subfigure}
% \hfill
\begin{subfigure}{.12\textwidth}
   \includegraphics[width=\textwidth]{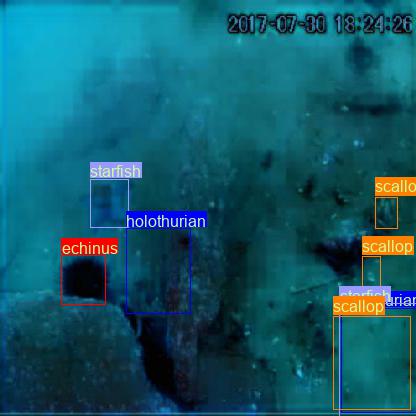}
   \caption{Model 2}
\end{subfigure}
% \hfill
\begin{subfigure}{.12\textwidth}
   \includegraphics[width=\textwidth]{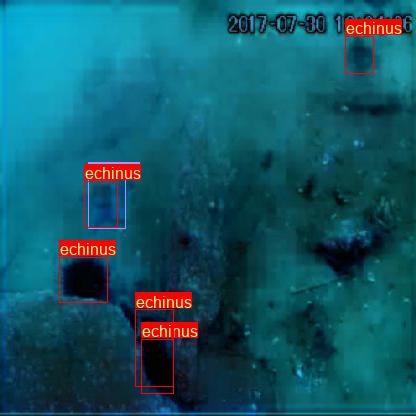}
   \caption{Model 3}
\end{subfigure}
% \hfill
\begin{subfigure}{.12\textwidth}
   \includegraphics[width=\textwidth]{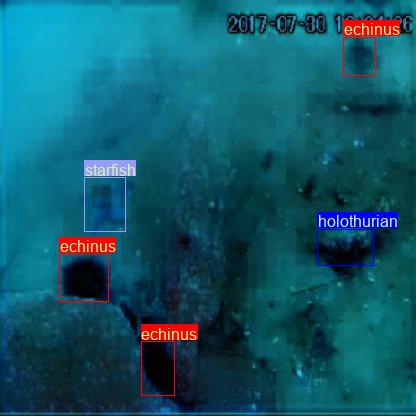}
   \caption{Model 4}
\end{subfigure}
% \hfill
\begin{subfigure}{.12\textwidth}
   \includegraphics[width=\textwidth]{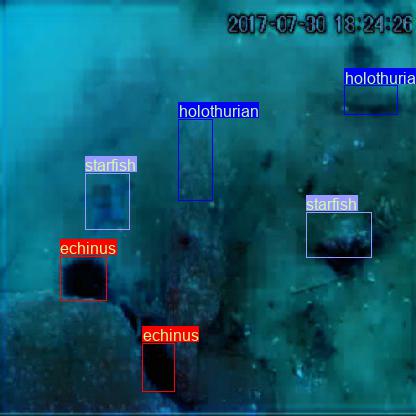}
   \caption{Model 5}
\end{subfigure}
% \hfill
\begin{subfigure}{.12\textwidth}
   \includegraphics[width=\textwidth]{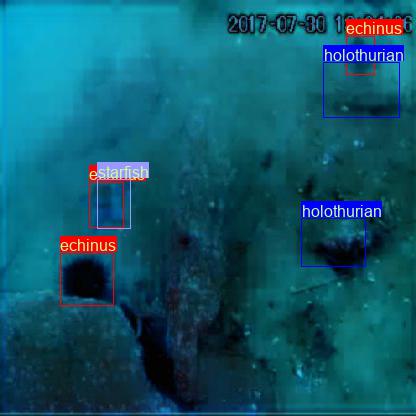}
   \caption{Model 6}
\end{subfigure}
% \hfill
\begin{subfigure}{.12\textwidth}
   \includegraphics[width=\textwidth]{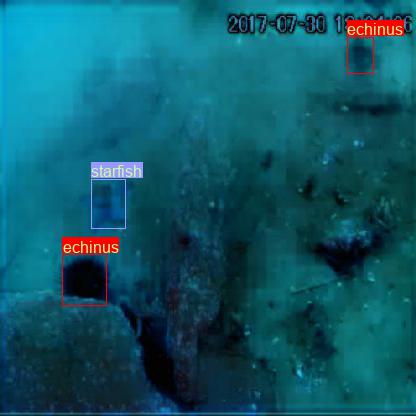}
   \caption{Model 7}
\end{subfigure}
% \hfill
\begin{subfigure}{.12\textwidth}
   \includegraphics[width=\textwidth]{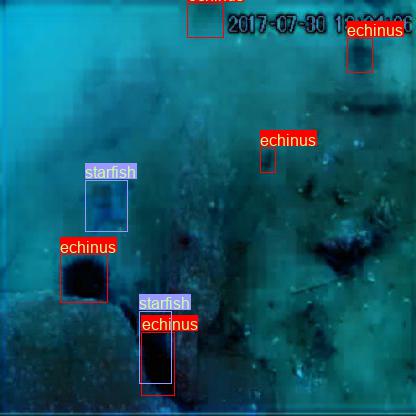}
   \caption{Model 8}
\end{subfigure}

\caption{Performance comparison for the obtained results using the original dataset (top) and the augmented dataset (Type 8) (bottom): The models from 1 to 8 are as follows: Baseline (YOLOv3), YOLO3+Residual, YOLO3+Channel attention, YOLO3+Residual attention, YOLO3+Domain, YOLO3+Domain+Residual, YOLO3+Domain+Channel attention, and YOLO3+Domain+Residual attention, respectively.}   
\label{fig:1}
\end{figure*}
We conducted an extensive evaluation of the proposed ADOD model on the publicly available dataset, Underwater Robot Picking Contest 2019 (URPC2019\footnote{Underwater Robot Professional Contest: http://en.urpc.org.cn}), which consists of 3765 training samples and 942 validation samples distributed among five categories: echinus, starfish, holothurian, scallop, and waterweeds. Our main objective was to conduct a comparative analysis of ADOD's performance for underwater object detection against the original YOLOv3 model. Table \ref{tab:results} presents the obtained results for each model.

During the experiments, we employed a high-performance workstation equipped with an NVIDIA GeForce RTX 4090 GPU and utilized the PyTorch 1.13.1 framework for model training. To ensure uniformity, all images were resized to a width of 416 pixels. For optimization, we used the Adam optimizer with a learning rate of $10^{-3}$, a batch size of 32, and conducted training for 300 epochs. These settings were consistently maintained across all models evaluated in this research.

The baseline YOLOv3 model achieved an overall mean Average Precision (mAP) of 54.28 on the original validation dataset and mAP of 33.01 on augmented validation dataset. While it performed reasonably well for echinus, starfish, holothurian, and scallop, it failed to detect waterweeds completely (mAP of 0.00). This deficiency suggests that the baseline model struggles with this particular class, possibly due to its distinctive appearance and variations in underwater conditions.

The addition of residual attention to the YOLOv3 model resulted in an overall mAP of 57.09 on the original validation dataset, demonstrating a slight improvement over the baseline. This enhancement was particularly effective for scallop, as it achieved the highest mAP of 69.12 among all models. However, similar to the baseline, it failed to detect waterweeds.

The YOLO3+Domain model showed competitive results, especially for starfish (mAP of 77.66 on the original validation dataset) and holothurian (mAP of 59.86). However, its performance for other classes remained similar to the baseline. The results suggest that domain adaptation had a positive impact on these two classes but did not significantly improve detection for the other categories.

The combination of domain adaptation and residual attention mechanisms led to a balanced performance across multiple classes, achieving an overall mAP of 35.90 on the augmented validation dataset. This model excelled in detecting starfish (mAP of 38.91) and showed improved performance for holothurian (mAP of 34.28). The introduction of residual attention likely contributed to enhanced detection for waterweeds, while domain adaptation played a role in improving detection for starfish.

The YOLO3+Domain +Channel attention model showed competitive results, with an overall mAP of 56.21 on the original validation dataset. However, it did not show significant improvements for any specific class compared to the baseline or other models.

Overall, the models with domain adaptation showed promising results for certain classes (starfish and holothurian), indicating that domain generalization can enhance the detection performance for these categories in various underwater environments. However, the impact of domain adaptation on other classes was limited.

Fig. \ref{fig:1} compares the performance of different models using two datasets: the original dataset (top) and the augmented dataset (Type 8) (bottom). The models include variations of YOLOv3 with different configurations, such as residual connections, channel attention, and domain adaptation techniques. The results demonstrate that domain-aware model 8 (YOLO3+Domain +Residual attention) outperform the baseline and other configurations on the augmented dataset, showcasing their ability to adapt to diverse underwater scenarios and improve object detection accuracy. This highlights the significance of domain generalization techniques for robust underwater object detection in various real-world applications.

% The evaluation and analysis of ADOD's performance demonstrate its potential as an accurate and robust model for underwater object detection, with improvements over the original YOLOv3 in certain object classes. These findings hold important implications for marine applications, including ecological monitoring, underwater exploration, and resource management.

The ADOD model exhibits promising results in underwater object detection, yet there are several limitations that require careful consideration:
\begin{itemize}

\item Dataset Bias: During evaluation, ADOD was tested on the URPC2019 dataset, which may introduce biases. As a result, further investigation is necessary to assess the model's generalizability to diverse underwater datasets. Future research should involve testing on larger and more varied datasets to enhance the model's robustness across different underwater environments.

\item Limited Detection of New Classes: ADOD demonstrates exceptional performance in recognizing specific underwater objects, such as echinus, starfish, holothurian, scallops, and waterweeds. However, its effectiveness in detecting new classes or other underwater items needs further exploration. To broaden its detection capabilities, efforts should be made to improve the model's ability to generalize to novel classes.

\item Resource Intensive: The integration of residual layers and multiscale attention in ADOD demands significant computational resources, making it challenging for deployment on resource-constrained or real-time systems. Enhancing the model's computational efficiency is vital to ensure its practicality for real-world applications.

\item Hyperparameter Sensitivity: ADOD's performance is sensitive to hyperparameters, including learning rate, batch size, and image size. Proper optimization of these hyperparameters across datasets and settings is crucial to achieve optimal performance in diverse underwater scenarios.

\item  Comprehensive Evaluation Metrics: While mean Average Precision (mAP) provides a comprehensive evaluation of object identification, other aspects such as localization accuracy and resilience to occlusions may not be fully addressed. Exploring additional evaluation metrics would offer a more comprehensive assessment of the model's strengths and limitations.
\end{itemize}

To address these challenges, future work should involve exploring various YOLO variants, such as YOLOv7 and YOLOv8, to evaluate their accuracy, speed, and robustness in underwater conditions. By addressing these limitations, we can advance underwater object detection and enhance the practical applicability of the ADOD model in real-world scenarios.

\section{CONCLUSIONS}

In this research, we introduced ADOD, an innovative underwater object detection approach based on YOLOv3, shows promising results. With modifications like residual layers and channel attention modules, ADOD significantly enhances detection performance in challenging underwater conditions.
On the original URPC2019 dataset, ADOD achieves an impressive mAP of 57.09\% compared to the baseline's 54.28\%. It effectively detects echinus, starfish, holothurian, and scallop.
Moreover, ADOD addresses domain shift challenges on the augmented type 8 dataset, achieving an mAP of 35.90\% compared to the baseline's 33.01\%. This highlights its robustness in detecting objects in unseen underwater conditions.
ADOD's enhanced detection capabilities have valuable implications for underwater robotics, marine biology research, and environmental monitoring.
Future work includes optimizing ADOD for real-time performance and evaluating it on diverse datasets to enhance adaptability. ADOD marks a significant milestone in underwater object detection.

\section*{Acknowledgement}
\noindent This work is supported by the Khalifa University of Science and Technology under Award No.  CIRA-2021-085,  RC1-2018-KUCARS.

%%%%%%%%%%%%%%%%%%%%%%%%%%%%%%%%%%%%%%%%%%%%%%%%%%%%%%%%%%%%%%%%%%%%%%%%%%%%%%%%

\bibliographystyle{IEEEtran}
\bibliography{IEEEabrv,references}

\end{document}